\documentclass{article}

\usepackage[preprint]{neurips_2023}

\usepackage[utf8]{inputenc} 
\usepackage[T1]{fontenc}    
\usepackage{hyperref}       
\usepackage{url}            
\usepackage{booktabs}       
\usepackage{amsfonts}       
\usepackage{nicefrac}       
\usepackage{microtype}      
\usepackage{xcolor}         

\usepackage{multirow}
\usepackage{graphicx}
\usepackage{subcaption}
\graphicspath{{figures/}} 
\usepackage{mathrsfs,amsmath}
\usepackage{bbm}

\title{SHAPNN: Shapley Value Regularized Tabular Neural Network}

\author{%
  Qisen Cheng* \\
  Samsung Display America Lab\\
  San Jose, CA 950134 \\
  \texttt{qisen.c@samsung.com} \\
  \And
  Shuhui Qu* \\
  Samsung Display America Lab \\
  San Jose, CA 950134 \\
  \texttt{shuhui.qu@samsung.com} \\
  \AND
  Janghwan Lee \\
  Samsung Display America Lab \\
  San Jose, CA 950134 \\
  \texttt{jake.ee@samsung.com} \\
}

\begin{document}

\maketitle
\def\thefootnote{*}\footnotetext{These authors contributed equally to this work}\def\thefootnote{\arabic{footnote}}

\begin{abstract}
We present SHAPNN, a novel deep tabular data modeling architecture designed for supervised learning. Our approach leverages Shapley values, a well-established technique for explaining black-box models. Our neural network is trained using standard backward propagation optimization methods, and is regularized with real-time estimated Shapley values. Our method offers several advantages, including the ability to provide valid explanations with no computational overhead for data instances and datasets. Additionally, prediction with explanation serves as a regularizer, which improves the model's performance. Moreover, the regularized prediction enhances the model's capability for continual learning. We evaluate our method on various publicly available datasets and compare it with state-of-the-art deep neural network models, demonstrating the superior performance of SHAPNN in terms of AUROC, transparency, as well as robustness to streaming data.
\end{abstract}

\section{Introduction}
Tabular data is widely used in real-world applications like scientific analysis [\cite{kehrer2012visualization}], financial transactions [\cite{andriosopoulos2019computational}], industrial planning [\cite{hecklau2016holistic}], etc. Tabular data are commonly presented in a structured and heterogeneous form [\cite{borisov2022deep}], with data points or samples in rows, and features in columns, corresponding to particular dimensions of information. 

In the past decade, machine learning algorithms have been used to efficiently analyze tabular data, with most research focusing on classification and regression tasks [\cite{athmaja2017survey}]. Gradient-boosted decision trees (GBDT) [\cite{chen2016xgboost}] and its extensions, such as LightGBM [\cite{ke2017lightgbm}] and CatBoost [\cite{dorogush2018catboost}], have emerged as dominant methods. However, these methods have limitations in practice due to their data-specific learning paradigm [\cite{arik2021tabnet}]. Firstly, gradient-based tree structures impede continual learning, which is crucial in situations where live data streams in. Secondly, these models are typically data-specific and must be learned in a fully supervised manner, which hinders their ability to fuse with other models and data modalities under different degrees of label availability [\cite{ke2019deepgbm}].


Recently, deep learning has been explored as an alternative to GBDT-based models for analyzing tabular data [\cite{huang2020tabtransformer}]. DNN employs adaptable weights that can be gradually updated to learn almost any mapping from inputs to targets, and it has proven to be effective and flexible in handling various types of data modalities. DNN models can also conveniently learn from and adapt to continuously streaming data [\cite{ke2019deepgbm}]. However, despite these promising features, DNN's performance on tabular data often falls short compared to that of GBDT-based methods [\cite{gorishniy2021revisiting}]. Additionally, DNN models are often considered a "black box" approach, lacking transparency in how they transform input data into model outputs [\cite{klambauer2017self}]. Due to these limitations of both GBDT and DNN, there is no clear winner for tabular data tasks [\cite{kadra2021well}, \cite{shwartz2022tabular}]. In comparison to GBDT-based models, DNN lacks two crucial capabilities, which degrade its performance on various tabular tasks: (1) the ability to effectively utilize the most informative features through the splitting mechanism based on information gain and (2) the capacity to progressively discover feature sets that lead to fine-grained enhancements through the boosting-based ensemble. We could contend that both capabilities contribute to evaluating feature utility and selecting relevant features during model training [\cite{grinsztajn2022tree}]. 


In this study, we aim to address these challenges faced by current deep learning methods for tabular data. Our objective is to develop a DNN-based model that accomplishes the following goals: (i) achieves superior performance on tabular data tasks, (ii) provides quantitative explanations of model decisions, and (iii) facilitates effective continual learning. To achieve these goals, we introduce SHAPNN, which leverages the Shapley value as a bridge between GBDTs and DNNs. The Shapley value is a model-agnostic approach used for generating post-hoc explanations by quantifying the influence of each feature on predictions based on game theory. In SHAPNN, we incorporate Shapley value estimation into the DNN training process and use it as additional supervision to transfer feature evaluation and selection guidelines from GBDT-based priors. However, Shapley value estimation is time-consuming due to exponentially growing feature selections [\cite{lundberg2017unified}]. To overcome this obstacle, we utilize the recent FastSHAP framework (\cite{jethani2021FastSHAP}) to efficiently estimate Shapley values and generate model predictions in a single forward propagation. Our approach also allows us to ensemble multiple prior models to provide comprehensive feature evaluation and selection guidelines. Moreover, in inference, we utilize the estimated Shapley values to obtain feature-level explanations of how the model makes decisions. We extend the utilization of Shapley values to enhance the continual learning of DNNs by using them as proxies that memorize the mapping from features to predictions in a certain time step. We can then use them to regulate the updating of models to achieve overall stability, eliminating the need for collecting and accessing all historical data during inference. Our extensive experiments demonstrate the effectiveness of the SHAPNN approach. Our contributions are threefold: 1) To our best knowledge, this is the first work to incorporate Shapley value estimation in DNN training for tabular data, 2) We demonstrate that the approach can improve overall stability in continual learning of DNN, 3) this method can be applied to different backbone models, resulting in performance improvements and quantitative explanations in a single feedforward pass.

In this paper, our motivations are introduced in section 2. The background of Shapley values is presented in section 3. Our proposed methodology is shown in section 4. The experiment details and results are presented in section 5. Related work is shown in section 6. Section 7 concludes our paper.

\section{An empirical study on tabular data}
\begin{figure}[h]
\vspace{-0.1in}
\begin{subfigure}{0.48\textwidth}
     \centering
     \includegraphics[width=1.\textwidth]{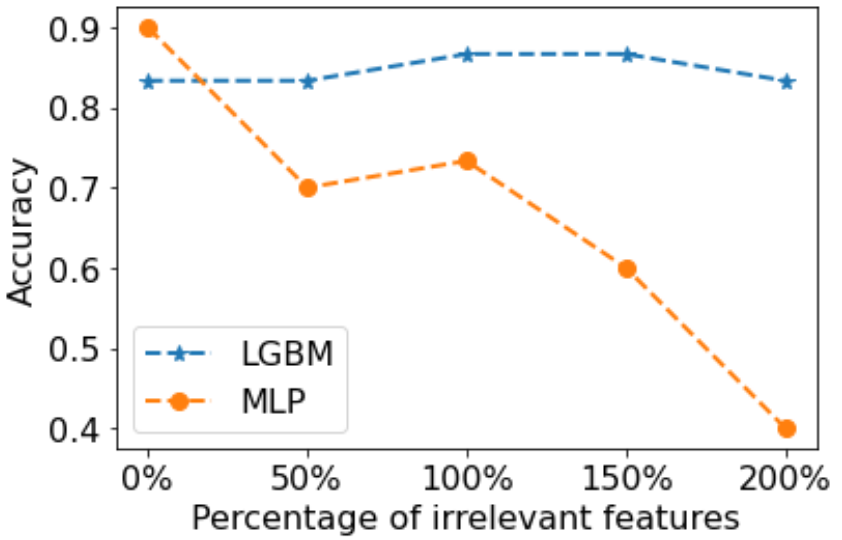}
     \caption{}\label{empirical_perf}
   \end{subfigure}\hfill
   \begin{subfigure}{0.48\textwidth}
     \centering
     \includegraphics[width=1.\textwidth]{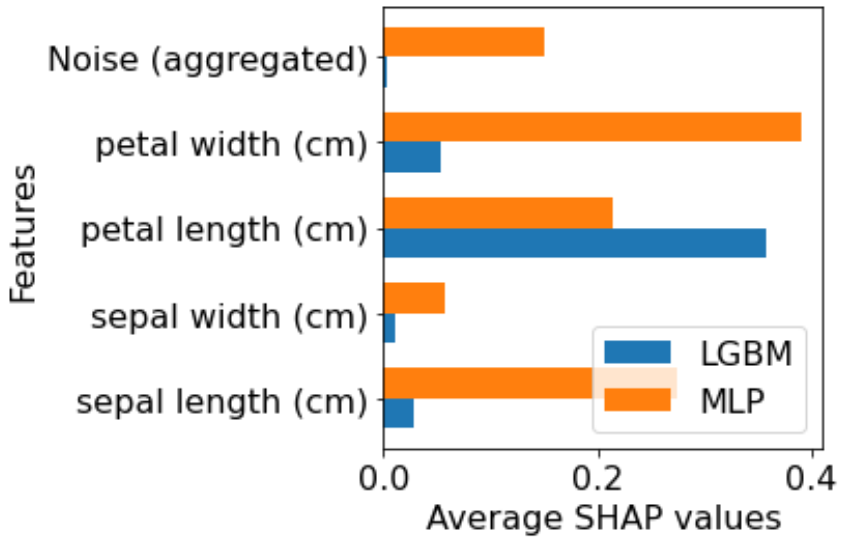}
     \caption{} \label{empirical_shap}
   \end{subfigure}
   \caption{Comparison between LGBM and MLP under impact of noisy features. In (a), it shows the prediction accuracy comparison between the 2 models with different amount of noisy features. (b) shows the Shapley values of each feature for both models.}
\end{figure}

We provide an empirical example to explain our motivation for introducing Shapley-based regularization into deep neural networks (DNN) training for tabular data. This example illustrates the shortcomings of using a Multilayer Perceptron (MLP) for feature evaluation and selection compared to a Gradient Boosting Decision Tree (GBDT) model. We compared the classification accuracy of LGBM (GBDT-based model) and MLP on a customized Iris dataset [\cite{fisher1936use}], where we purposefully attach extra numerical features (columns) whose values are sampled from a uniform distribution. As demonstrated in Figure \ref{empirical_perf}, we observe a significant decrease in MLP's classification accuracy, as the percentage of extra features being constructed. We further investigate the effect of each feature on model prediction by examining their Shapley values by using KernelSHAP [\cite{lundberg2017unified}]. As shown in Figure \ref{empirical_shap}, we observe that the extra features obtain a larger impact on MLP predictions, which explains its performance degradation. In contrast, the GBDT model almost completely disregards the extra features, and its performance remains stable even with the introduction of the new features.

The aforementioned example suggests a potential remedy for the comparatively weaker feature evaluation and selection ability of DNNs. As Shapley values provide a measure of the contribution of each feature, we can align the values obtained by DNNs with those obtained by GBDTs, in order to supervise the training process. This approach has the potential to enhance DNNs training by reducing the impact of irrelevant features and prioritizing the learning of useful ones.

\section{Background}
\subsection{Shapley value}
The Shapley value aims to distribute the gain and cost fairly among the players in a coalition game to achieve a desired outcome or payoff. In a coalition game, there are $N$ players and a characteristic function $v(S)$ that maps the subset of players $S\in\{0,1\}^N$ to a real number, representing the expected sum of the payoffs of the subset S that can be obtained through cooperation. The Shapley value $\phi(v)\in R^N$ distributes the gains to players. The contribution $\phi_i(v)$ of player $i$ is calculated:
\begin{equation}
\phi_i(v) = \frac{1}{N} \sum_{S\subseteq N \backslash {i}} \binom{n-1}{|S|}^{-1} \left(v(S\cup{i}) - v(S)\right)
\end{equation}

In the context of machine learning explanation, the characteristic function shows how the prediction of a sample changes when different subsets of features are removed. More specifically, given a sample $(x,y)\in \mathcal{D}$ from dataset $\mathcal{D}$, where $x=(x_1,...,x_N)$ is the input vector, and $y\in{1,...,K}$ is the output of $K$ classes for a classification problem, the characteristic function $v$ is defined as follows:
\begin{equation}
v_{x,y}(S)=E_{p(x_{\mathbbm{1}-S}})[\operatorname{Softmax}(f_{\theta}(x_S,x_{\mathbbm{1}-S})], y)
\end{equation}
Here, $f_{\theta}$ represents the machine learning model.The exact computation of the Shapley value increases exponentially with the number of players (features) $N$. Various approximation solutions have been proposed to improve efficiency [\cite{lundberg2017unified}]. Despite these, accurately estimating the Shapley value can still be extremely slow for large-scale and high-dimensional cases.

\subsection{FastSHAP}
Due to the computational cost of Shapley value estimation, we adopt the FastSHAP approach introduced in [\cite{jethani2021FastSHAP}] to perform amortized estimation of Shapley values. Specifically, we first learn a FastSHAP function $\phi_{fast,\gamma}(x,y): X\times Y \rightarrow R^N$, with $\gamma$ being the Shapley value generation model that map each feature to a Shapley value. The function is learned in a single forward pass by penalizing predictions using the following loss:
\begin{equation}
\begin{split}
    \mathcal{L}_{\gamma} = E_{p(x)} E_{U(y)} E_{p(S)}[(v_{x,y}(S) - v_{x,y}(\emptyset) -  S^T\phi_{fast,\gamma}(x,y))^2]
\end{split}
\end{equation}

where $S$ denotes a subset of features, $p(S) = \frac{n-1}{\binom{n}{\mathbbm{1}^TS} \mathbbm{1}^T S (N - \mathbbm{1}^TS)}$, and $U(y)$ is uniformly sampled from a distribution of $K$ classes. To further improve training efficiency, we use additive efficient normalization to obtain Shapley value estimation function $\phi_{fast,\gamma}^{eff}(x,y)$:
\begin{equation}
\begin{split}
    \phi_{fast,\gamma}^{eff}(x,y) =  &\phi_{fast}(x,y,\gamma) + \frac{1}{N} \left(v_{x,y}(\mathbbm{1}) - v_{x,y}(\emptyset) - \mathbbm{1}^T  \phi_{fast,\gamma}(x,y)\right)
\end{split}
\end{equation}
Here, $\phi_{fast,\gamma}(x,y)$ denotes the original FastSHAP function, $v_{x,y}(\mathbbm{1})$ and $v_{x,y}(\emptyset)$ are the sum of prediction values of all features and the sum of prediction values with zero features, respectively. FastSHAP consists of three steps: 1) Train a machine learning model $f_{\theta}(x)\rightarrow y$ to be explained. 2) Train a surrogate model $f_{surr,\beta}$ that approximates the original prediction model $f_{\beta}(x, m)\rightarrow y$ with a masking function $m(x,S)$ that is to replace the feature $x_i$ with a default value not in the support of $S$. 3) Train the Shapley value generation model $\phi_{\gamma}(x)\rightarrow v_{x}(S)$. 


\begin{figure}[h]
\vspace{.3in}
\centerline{\includegraphics[width=1.\textwidth]{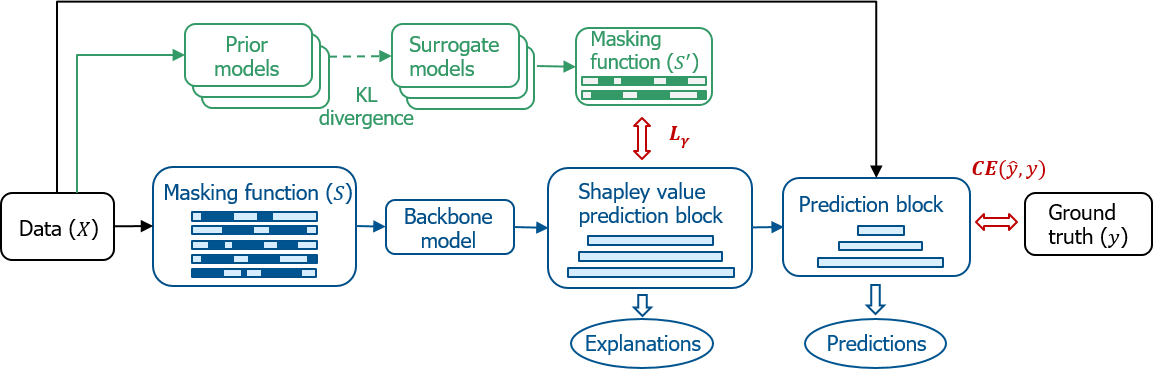}}
\caption{The SHAPNN framework involves generating Surrogate Models from prior models. During the training of the DNN, the input data $X$ is perturbed, and the Shapley values for the perturbed dimensions are estimated in an intermediate block. The original data and the estimated Shapley values are then passed to the prediction block for classification.}
\end{figure}

\section{Methodology}
\subsection{SHAPNN}
This section presents the Shapley-based Neural Network (SHAPNN), which is built upon FastSHAP. By utilizing estimated Shapley values as intermediate features, SHAPNN is designed to construct a machine-learning prediction model that achieves both high prediction accuracy and interpretability. Both model predictions and Shapley value estimations are obtained in a single forward pass.

The neural network serves as the foundation for (1) the Shapley value generation model $\phi_{\gamma}(x)\in R^{N\times K}$, which takes the input feature vector and generates the Shapley value vector for each possible class, and (2) the surrogate model $f_{\beta}(x,S) \rightarrow y$, which takes the input feature vector and support $S$ to produce the predicted label. 

The Concat SHAPNN $f_w(x)$ is constructed by incorporating the estimated Shapley value $v_{x}(S)$ as part of the input feature to the prediction model $f_{w'}: f_{w'}(f_{w\setminus w'}(x),v_{x}(S))\rightarrow y$, where $v_{x}(S)=\phi_{\gamma}(x) \in R^{N,K}$ represents the estimated Shapley value. The Concat SHAPNN's loss function ($\mathcal{L}$) is:
\begin{equation}
\mathcal{L} = \mathcal{L}_{\gamma} + CE(y',y)
\end{equation}
Here, $CE$ denotes the cross-entropy loss function for classification.



\subsection{Ensemble prior}
The SHAPNN with ensemble prior is developed by aligning estimated Shapley values to a series of GBDT models, such as an ensemble prior that combines Xgboost and LightGBM. These Shapley value estimations are then integrated into the input feature for the prediction model, $f_{w'}(f_{w\setminus w'},v_{x}(S))\rightarrow y'$, where $v_{x}(S)=\phi_{\gamma}(x)\in R^{N,K}$ is the estimated Shapley value. The overall SHAPNN loss function is defined as:
\begin{equation}
\mathcal{L} = agg_k(\mathcal{L}_{\gamma}^{k}) + CE(y',y)
\end{equation}
where the aggregation operator, $agg$, combines the losses from each prior model of the ensemble, indexed by $k$. In practice, we use a weighted sum for aggregation. This design of the SHAPNN enables explainability while also achieving higher performance.


\subsection{Continual learning}
The concept of continual learning can be framed as follows: given a data stream composed of a series of data batches ${x^t}$, indexed by $t\in [0, 1, ... ,T]$, and a model $f_{w}(x)\rightarrow y$ that is sequentially trained on each data batch $x^t$ and recorded at each time step as ${f_{w}^t}$, the task is to make two predictions at each time step. Firstly, using the most recent recorded model ($f_{w}^{t-1}$), we make predictions ($\hat{{y}^t}$) on the current data batch $x^t$. Note that this batch of data is not available for model training before making the prediction. Secondly, we make backward predictions ($\tilde{{y}^{t-1}}$) on data batches that precede $t-1$ using $f_{w}^{t-1}$. Our aim is to ensure that both $\hat{{y}^t}$ and $\tilde{{y}^{t-1}}$ are accurate predictions of their respective true labels $y$.

During each time step $t$, the model is trained using a combination of model prediction loss and Shapley estimation regularization, as described in previous sections. To ensure the model remains robust to concept drift, we generate pseudo labels for time step $t$ by applying mixup [\cite{zhang2017mixup}] between the true label and all predictions from surrogate models of previous steps. This involves combining the true label ($y^t$) with a weighted average of the predictions ($f_{w}^{i}(x^t)$) from previous steps $i\in \{1,...,t-1\}$, where the weight is controlled by a parameter $\alpha$:
\begin{equation}
\widetilde{{y}^{t}} = \alpha\cdot{y^{t}} + (1-\alpha)\cdot\sum_{i}^{t-1}{f_{w}^{i}(x^t)}
\end{equation}
To ensure stable feature selection and evaluation during continual learning, we also extend the regularization by including all the explanation models $\gamma_t$ from previous time steps. Thus, the model at time step $t$ is trained using the following loss:
\begin{equation}
\mathcal{L}^{t} = \sum_{i}^{t-1}\lambda^{i}\cdot\mathcal{L}_{\gamma}^{i} + CE(\hat{{y}^t},\widetilde{{y}^{t}})
\end{equation}
where $\lambda_i$ is a discount factor of the losses from each time step, and $\sum_{i}^{t-1}\lambda^{i} = 1$. In practice, we use a decaying schedule that emphasizes recent steps and reduces the effect of distant steps.

\section{Experiments}
\subsection{Implementation and setup}
To evaluate the generalizability of our SHAPNN approach, we conducted our experiments on two popular DNN models for processing tabular data: Multi-Layer Perceptron [\cite{kadra2021well}] and recently published FT-Transformer [\cite{gorishniy2021revisiting}], which has demonstrated state-of-the-art performance on various tabular datasets. The MLP has 3 hidden layers, each containing 512 neurons, while the FT-Transformer's hyperparameter follows [\cite{gorishniy2021revisiting}]. The Shapley estimation block for both implementations consists of a 2-layer MLP with an output dimension equal to the number of features in each dataset. The prediction layer is a linear projection layer without nonlinear activation functions. We employed the standard Stochastic Gradient Descent (SGD) optimizer and followed the hyper-parameter settings outlined in [\cite{gorishniy2021revisiting}], including the learning rate selection. More detail is shown in Appendix.

\subsection{Tabular data analysis and datasets}
We conducted experiments on several well-known benchmark datasets, including: 1) the Adult Income dataset [\cite{kohavi1996scaling}], which comprises 48842 instances of adult income data with 14 attributes; 2) the Electricity dataset [\cite{hoiem2009pascal}], which contains 45312 instances of electricity consumption with 8 real-valued attributes; 3) the Iris dataset [\cite{fisher1936use}], consisting of 3 types of Iris flowers, each with 50 samples; 4) the Epsilon dataset [\cite{blackard1999comparative}], comprising 400000 objects with 2001 columns of simulated experiments; and 5) the Covertype dataset [\cite{hulten2001mining}], which includes 581012 instances of tree samples, each with 54 attributes. We specifically chose the Epsilon and Covertype datasets for their higher dimensionality, which allowed us to demonstrate the efficiency and scalability of our method. The evaluation metric used for all analyses in this section is the Area Under the Receiver Operating Characteristic curve (AUROC). We chose this metric to ensure a fair comparison and to account for label imbalance bias.

\begin{table}[]
\centering
\caption{Prediction Results (AUROC)} \label{tabperf}
\begin{tabular}{@{}lllllll@{}}
\toprule
Datasets &                      & Adult & Electricity & Iris  & Epsilon & Covertype \\ \midrule
\multirow{6}{*}{Models} & Logistic Regression        & 0.793        & 0.774       & 0.935 & 0.854   & 0.945     \\
                        & Random Forrest          & 0.837        & 0.822       & 0.959 & 0.892   & 0.957     \\ \cmidrule(l){2-7}
                        & MLP                     & 0.839        & 0.790       & 0.946 & 0.883   & 0.955     \\
                        & \textbf{SHAPNN (MLP)}            & \textbf{0.852} & \textbf{0.818} & \textbf{0.952} & \textbf{0.892} & \textbf{0.961} \\ \cmidrule(l){2-7}
                        & FT-Transformer                   & 0.849          & 0.824          & 0.954          & 0.890          & 0.960          \\
                        & \textbf{SHAPNN (FT-Transformer)} & \textbf{0.857} & \textbf{0.835} & \textbf{0.957} & \textbf{0.894} & \textbf{0.969} \\ 
                        \bottomrule
\end{tabular}
\centering
\end{table}



\subsection{Model prediction results}
Table \ref{tabperf} shows that our SHAPNN approach applied to MLP consistently improves performance over the vanilla MLP baseline on all tabular data benchmarks. The magnitude of improvement appears to be associated with the difficulty of the datasets. On the challenging Adult Income dataset, which has missing values and different data types in features \cite{shwartz2022tabular}, we achieve an improvement in AUROC of 1.3\%. We observe an 0.6\% increase in AUROC over the original 94.6\% on the Iris dataset, which has the smallest size and fewest features among the five datasets. 


Table \ref{tabperf} also shows the test results on the FT-Transformer backbone, where we also observe improvements over the baseline model on all 5 test cases. Notably, FT-Transformer is a stronger baseline compared to MLP, potentially due to its attention mechanism that effectively weighs the features based on their pairwise correlation. Nevertheless, our approach still benefits FT-Transformer by enhancing feature evaluation and selection. Additionally, we compare the performance of two widely used models, Logistic Regression (LR) and Random Forest (RF), for tabular classification tasks to further evaluate our FT-Transformer's performance. The results show that FT-Transformer's performance is comparable to, or better than, that of LR and RF.

\begin{table}[]
\centering
\caption{Performance comparison between single and ensemble prior (AUROC)} \label{singlevsensemble}
\begin{tabular}{@{}cllllll@{}}
\toprule
Dataset &                       & Adult          & Electricity    & Iris  & Epsilon        & Covertype \\ \midrule
\multirow{2}{*}{Models} & SHAPNN (single prior)    & 0.849          & 0.807          & 0.952 & 0.889          & 0.961     \\
                        & SHAPNN (ensemble prior) & \textbf{0.852} & \textbf{0.818} & 0.952 & \textbf{0.892} & 0.961     \\ 
\bottomrule
\end{tabular}
\centering
\end{table}

\begin{table}[]
\centering
\caption{Inference speed comparison} \label{inferencetime}
\begin{tabular}{@{}lll@{}}
\toprule
\multirow{2}{*}{Dataset} & \multicolumn{2}{c}{Models} \\ \cmidrule(l){2-3} 
                         & SHAPNN     & KernelSHAP    \\ \cmidrule(r){1-3}
Epsilon                  & \textbf{4.7 s}      & 34.9 s        \\ 
Covertype                & \textbf{0.8 s}      & 5.2 s         \\ \bottomrule
\end{tabular}
\centering
\end{table}


\subsubsection{Single prior vs. ensemble priors}
The performance comparison between a DNN trained with a single prior model and an ensemble of prior models is presented in Table \ref{singlevsensemble}. The results show that on 3 of the 5 datasets, including the more challenging Adult Income and Electricity datasets, using ensemble priors leads to better performance compared to using a single prior. However, on the Iris and Covertype datasets, where the original performance is already high, the performance of using ensemble priors is the same as using a single prior. The observed improvement in performance may be attributed to the ensemble priors providing a more comprehensive evaluation of features compared to a single prior.

\subsection{Model explanation results}

\begin{figure}[h]
\begin{subfigure}{0.48\textwidth}
     \centering
     \includegraphics[width=1.\linewidth]{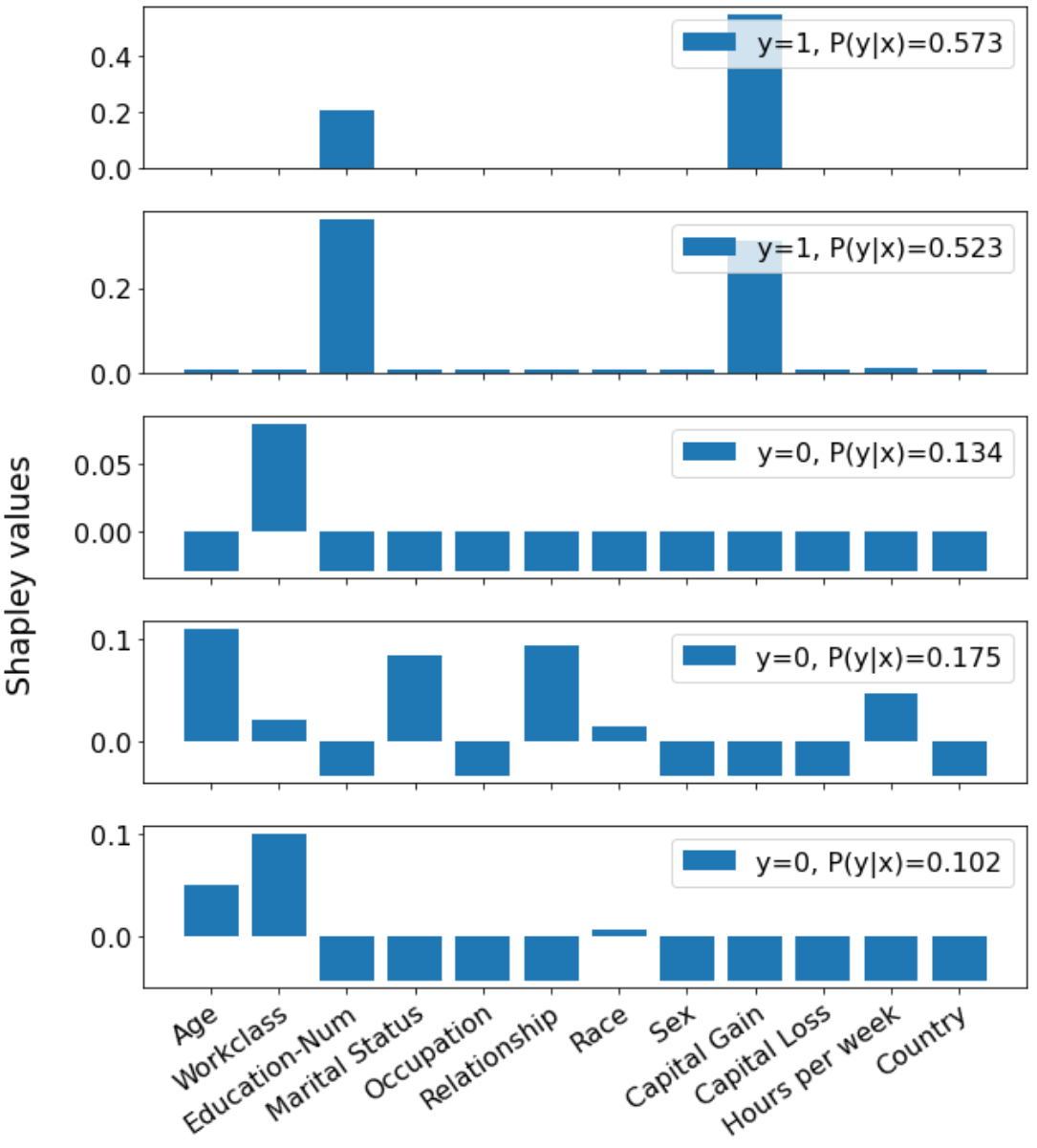}
     \caption{}\label{explain-sample-wise}
   \end{subfigure}\hfill
   \begin{subfigure}{0.48\textwidth}
     \vspace{0.55in}
     \centering
     \includegraphics[width=1.\linewidth]{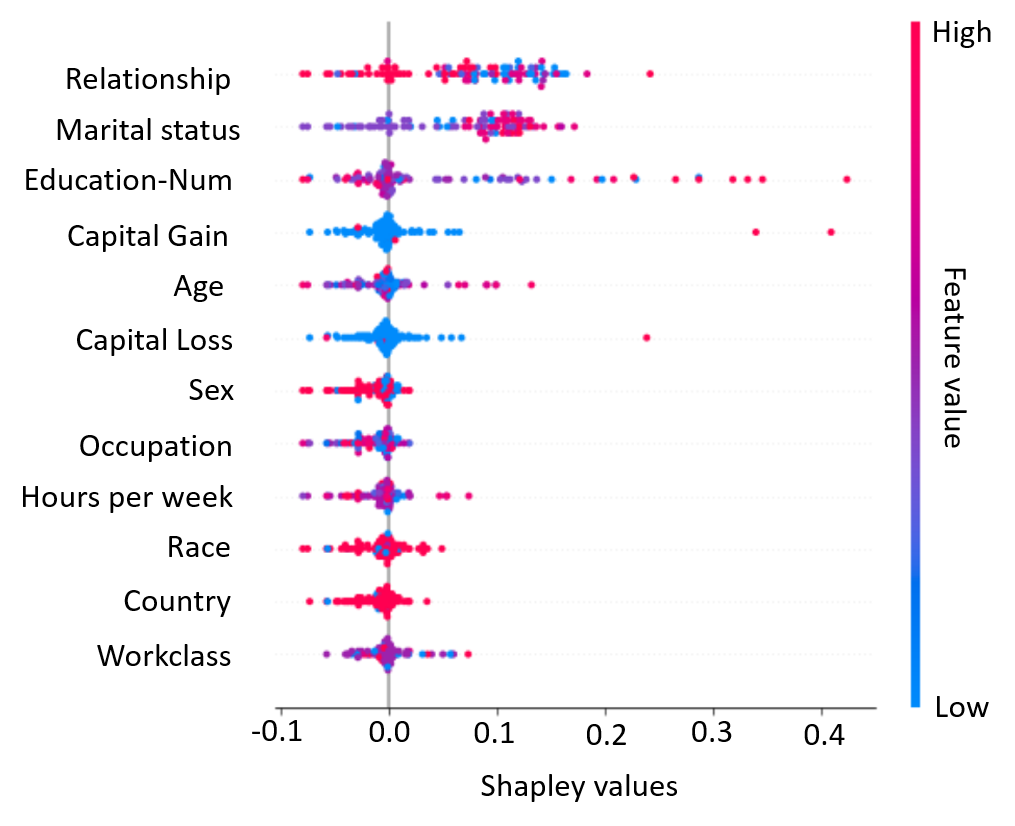}
     \caption{}\label{explain-population}
   \end{subfigure}
   \caption{Explanation examples of Adult Income Dataset. (a) shows the sample-wise explanation examples, and (b) gives the population-wise explanation examples.}
\end{figure}

Figures \ref{explain-sample-wise} and \ref{explain-population} illustrate SHAPNN's ability to provide quantitative explanations at both the sample-wise and population-wise levels, respectively, using the Adult Income dataset as an example. For each type of explanation, SHAPNN presents the impact of each feature on the model prediction, along with its magnitude and polarity.

In sample-wise explanations, the magnitude indicates the importance of each feature, while the polarity reflects the direction in which a feature influences the model prediction for a particular sample. For example, the education length feature seems to be important for predicting personal income, with a positive contribution to high earners and a negative contribution to low earners. Notably, negative class samples (i.e., low earners) are associated with features of overwhelmingly negative impacts, while positive class samples have more diverse feature influences.

Similarly, population-wise explanations demonstrate the general relationship between feature values and their influence within a given population. In this example, relationship and marital status are identified as two important factors. We can interpret from the plot that not being in a relationship or being married almost always contributes positively to earning status, whereas the influence is more diverse for opposite conditions. It is also worth mentioning that only a few features have high Shapley values, which could be an effect of the proposed regularization.

To evaluate the efficiency of our method in generating explanations, we conducted a wall-clock experiment comparing the inference time consumed by SHAPNN and KernelSHAP \cite{lundberg2017unified} for generating sample-wise explanations. We tested Covertype and Epsilon datasets due to their relatively higher dimensionality. We report the average inference time of 100 randomly sampled data points in Table \ref{inferencetime}. Our method was found to provide a 7-8X speedup over KernelSHAP.



\subsection{Continual learning analysis}
We further analyze the ability of SHAPNN in handling streaming data through the continual learning framework. Continual learning presents two conflicting challenges\cite{de2021continual}: the model should quickly adapt to incoming data that often leads to concept drift, but it should not forget the knowledge learned from previous data and become biased towards the newest data. To comprehensively evaluate the model's performance in both aspects, we conduct both online adaptations and retrospective tests. 

We use three synthetic streaming datasets with controlled levels of concept drift for this analysis: {STA dataset} \cite{gama2004learning}, {SEA dataset} \cite{street2001streaming}, and {ROT dataset} \cite{hulten2001mining}. In all three datasets, the mapping between features and predictors changes over time with different concept drifts defined by certain functions. Recurring and abrupt concept drift is introduced into each time window by randomly shuffling the parameter of the functions. The function definitions can be found in Appendix. These datasets pose a significant challenge to the model.


\subsubsection{Online adaptation}
For all the datasets that follow, we conduct an adaptation test by assuming that only the most recent data is available for re-training. This means that we test the model on each time step $t$ after updating it with the most recent data (i.e., data batch $t-1$). We compare two scenarios: one with SHAPNN and one without SHAPNN, using MLP as the backbone model (see Appendix) in both cases.

\begin{figure}[!htb]
\centering
   \begin{subfigure}{0.48\textwidth}
     \centering
     \includegraphics[width=\linewidth]{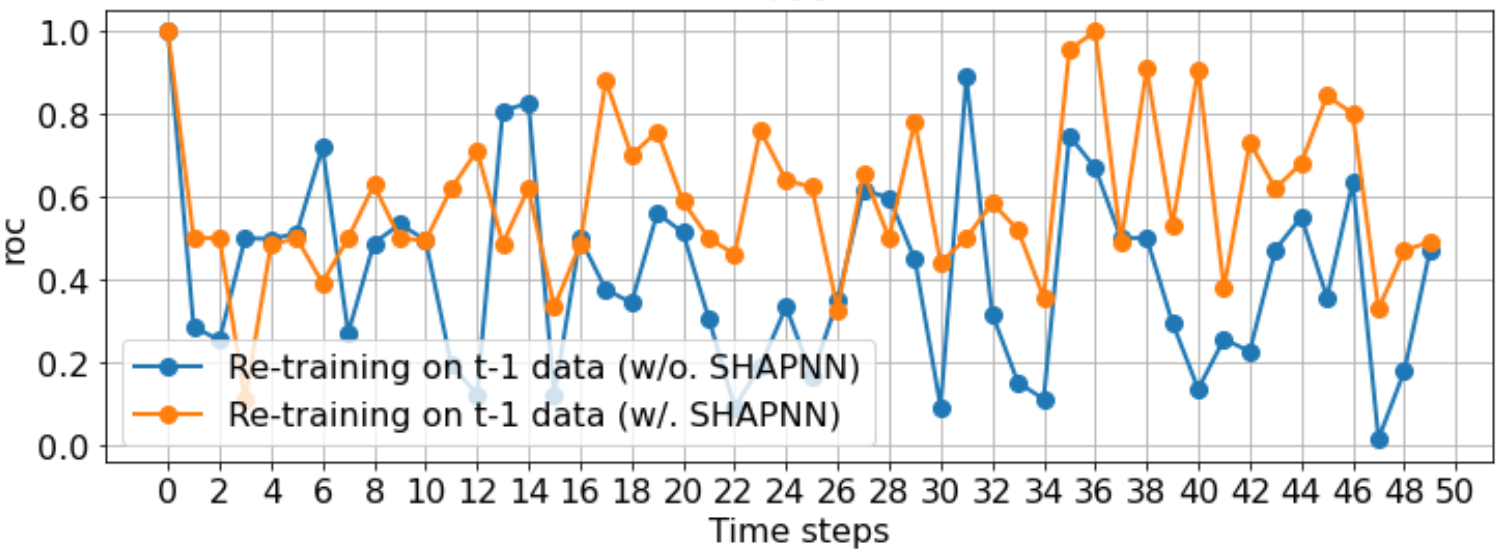}
     \caption{}\label{adapt_sta}
   \end{subfigure}\hfill
   \begin{subfigure}{0.48\textwidth}
     \centering
     \includegraphics[width=\linewidth]{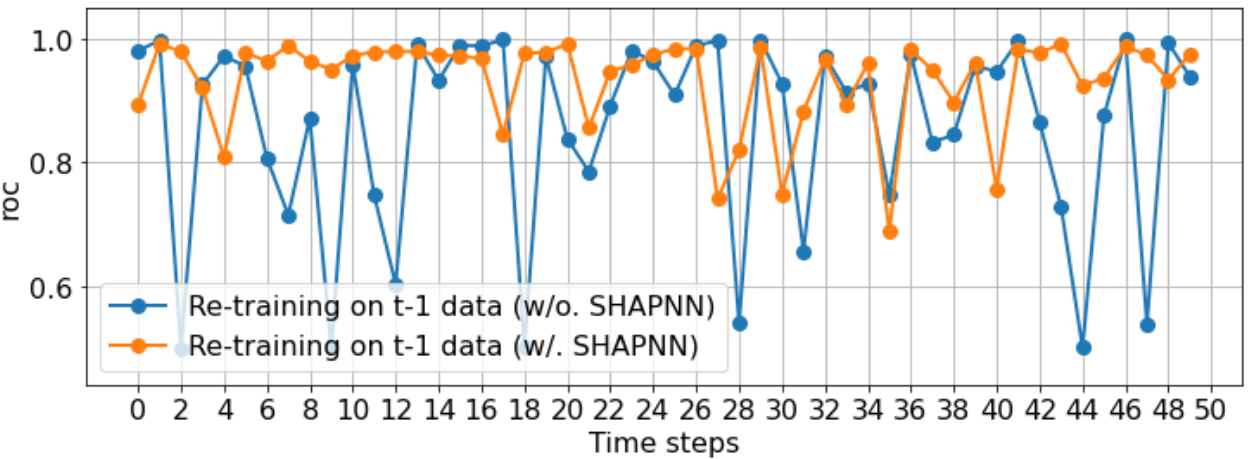}
     \caption{}\label{adapt_sea}
   \end{subfigure}
   \vskip\baselineskip
   \begin{subfigure}{0.48\textwidth}
     \centering
     \includegraphics[width=\linewidth]{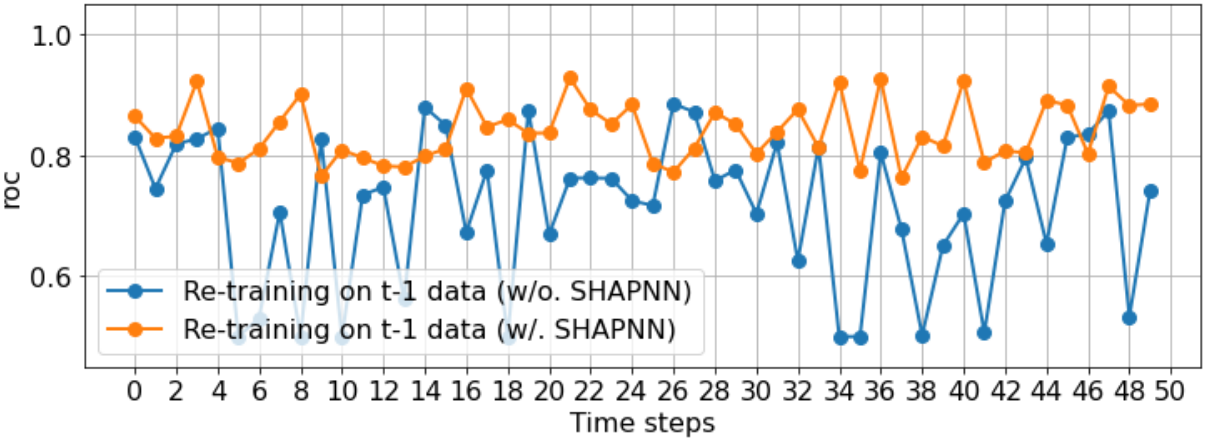}
     \caption{}\label{adapt_rot}
   \end{subfigure}
   \caption{Online adaptation performance for (a) STA, (b) SEA, and (c) ROT datasets.}
\centering
\end{figure}


Figures \ref{adapt_sta} to \ref{adapt_rot} depict the online adaptation results on these streaming datasets. The comparison between the baseline case and the SHAPNN approach reveals that the latter provides much more stable performance across all time steps. The fluctuations are reduced, and the average performance is substantially higher. These suggest SHAPNN's capability for online adaptation of streaming data.

\subsubsection{Retrospective test}
For this test, we update the MLP model (see Appendix) using the same approach as in the online adaptation test. We assess the model's performance by predicting the historic data it was trained from and report the average AUROC over all past time steps.


The retrospective testing outcomes are displayed in Table \ref{retrospective}. The test outcomes are reported at timestep 10 and 50. Since no historical data is used in model retraining, the MLP baseline model performs poorly on previous data batches after updating its weights at the evaluation time step. At timestep 50, the MLP model barely outperforms the random guessing, which clearly indicates the catastrophic forgetting issue. The model's weights are biased toward the latest data and lose previously learned concepts. On the other hand, SHAPNN consistently maintains a higher model performance on previous data batches, which shows the efficacy of SHAPNN in mitigating the catastrophic forgetting issue.

\begin{table}[]
\caption{Retrospective testing results (AUROC)}  \label{retrospective}
\centering
\begin{tabular}{@{}llllllll@{}}
\toprule
Dataset                 &              & \multicolumn{2}{l}{STA}         & \multicolumn{2}{l}{SEA}         & \multicolumn{2}{l}{ROT}         \\ \midrule
Timestep                &              & 10             & 50             & 10             & 50             & 10             & 50             \\ \midrule
\multirow{2}{*}{Models} & MLP          & 0.647          & 0.493          & 0.627          & 0.563          & 0.692          & 0.583          \\
                        & SHAPNN (MLP) & \textbf{0.715} & \textbf{0.673} & \textbf{0.902} & \textbf{0.757} & \textbf{0.881} & \textbf{0.785} \\ 
\bottomrule
\end{tabular}
\centering
\end{table}


\section{Related work}
\textbf{Neural networks for tabular data}
Several approaches have been proposed to enhance the performance of tree-based models for analyzing tabular data, either by extending them with deep learning techniques or by designing new neural architectures \cite{borisov2022deep}. Two main categories of model architectures have emerged from these efforts: differentiable trees and attention-based models. For instance, TabNet leverages sequential attention to perform feature selection and learning \cite{arik2021tabnet}, while NODE uses an ensemble of shallow neural nets connected in a tree fashion \cite{popov2019neural}. Another example is Net-NDF, which utilizes disjunctive normal neural form blocks to achieve feature splitting and selection \cite{katzir2020net}. More recently, researchers have explored applying transformer-based models to tabular data, with TabTransformer being the first attempt to do so \cite{huang2020tabtransformer}. This approach has been further improved upon in SAINT, which introduced additional row-wise attention \cite{somepalli2021saint}. The state-of-the-art method in this category is the Feature-tokenizer Transformer, which enhances the learning of embedding from tabular data with a tailored tokenizer \cite{gorishniy2021revisiting}.

\textbf{Intepretable machine learning}
The importance of generating interpretable tabular neural networks has gained increasing attention in recent years, particularly for critical applications where explanations are essential [\cite{sahakyan2021explainable}]. Existing work in this area often relies on attention-based mechanisms to generate feature-level explanations [\cite{konstantinov2022attention}]. Another line of research involves using model-agnostic approaches to explain trained models, such as KernelSHAP and its extensions [\cite{lundberg2017unified, covert2021improving}]. While most Shapley-based explanations are performed post-hoc, \cite{wang2021shapley} proposed a Shapley Explanation Network that incorporates Shapley values during training by adding extra Shapley value estimation modules to the neural net. In contrast, our approach uses amortized estimation to generate and leverage Shapley-based representations, which largely reduces the complexity of incorporating Shapley value.

\textbf{Continual learning}
Concept drift handling and adapting to new data after model training have been extensively discussed and explored even before the advent of deep learning [\cite{widmer1996learning, gama2014survey}]. Typically, existing work relies on collectively re-training a new model on the aggregated historical data. With deep learning, this concept has been extended to continual learning, which focuses on learning new tasks while preventing the model from forgetting what has been learned on old tasks [\cite{chen2018lifelong}]. As summarized in [\cite{de2021continual}], prior work has introduced more regularization terms during training [\cite{aljundi2018memory, zhang2020class}], learned separate sets of parameters for different tasks [\cite{aljundi2017expert, rosenfeld2018incremental}], or retained sampled historical data in a memory buffer to compensate for new task data during re-training [\cite{rolnick2019experience, lopez2017gradient}]. For instance, ASER [\cite{shim2021online}] leverages Shapley values to adversarially select buffered data samples for effective re-training. In a similar vein, we also utilize the Shapley values for continual learning. However, unlike ASER, we directly leverage the Shapley value estimator of past models as a medium for retaining knowledge from past training without accessing any historical data. Since the Shapley value estimators already contain the information on the mapping between features and predictions, we use them to regularize the parameter updating.

\section{Conclusion, Boarder Impact, Limitations, LLM Statement}

We introduce SHAPNN, a new deep-learning architecture for supervised learning tasks on tabular data. The neural network incorporates real-time Shapley value estimation module, which is trained through standard backward propagation. The estimation module provides enhanced regularization for model training that leads to performance improvements and enables valid explanations with no extra computational cost. Furthermore, the Shapley-based regularization improves the ability to perform continual learning. We extensively evaluate SHAPNN on publicly available datasets and compare it to state-of-the-art deep learning models, demonstrating its superior performance. We also show that SHAPNN is effective in continual learning, adapting to concept drifts and being robust to noisy data. 

Our work could potentially facilitate general data analysis, and improve the transparency and trustworthiness of AI. Some limitations of our method include: 1) prior models need to be trained separately, ahead of training of the neural network; 2) our model may have an upper limit on its capacity to adapt to new concepts or drifts. In this paper, we use LLM to correct grammatical mistakes.

\bibliography{SHAPNN_ref}
\renewcommand{\bibname}{References}
\renewcommand{\bibsection}{\subsubsection*{\bibname}}

\end{document}